\documentclass[10pt,twocolumn,letterpaper]{article}
\usepackage{authblk}
\usepackage[pagenumbers]{cvpr} 
\usepackage{pifont}
\usepackage{graphicx}
\usepackage{amsmath}
\usepackage{amssymb}
\usepackage{booktabs}
\usepackage{fullpage}
\usepackage{times}
\usepackage{fancyhdr,graphicx,amsmath,amssymb}
\usepackage[ruled,vlined]{algorithm2e}
\usepackage{multirow}
\usepackage[table,xcdraw]{xcolor}
\usepackage{float}
\usepackage{bbding}
\makeatletter
\@namedef{ver@everyshi.sty}{}
\makeatother
\usepackage{tikz}
\newcommand{\tikzcircle}[2][red,fill=red]{\tikz[baseline=-0.5ex]\draw[#1,radius=#2] (0,0) circle ;}%

\usepackage{colortbl}
\usepackage{diagbox}
\usepackage{rotating}
\usepackage{booktabs}
\usepackage{overpic}
\usepackage{enumitem}
\usepackage{colortbl}
\usepackage{algpseudocode}
\usepackage{cite}
\usepackage[export]{adjustbox}
\usepackage{listings}
\usepackage{pifont}
\usepackage{subcaption}
\usepackage{makecell}
\usepackage{arydshln}
\usepackage{tabu}
\usepackage{utfsym}

\definecolor{mygray}{gray}{.92}

\definecolor{mygray}{gray}{.9}
\definecolor{ggray}{RGB}{127,127,127}
\definecolor{reda}{RGB}{192,0,0}
\definecolor{redb}{RGB}{217,148,143}
\definecolor{myyellow}{RGB}{190,144,0}
\definecolor{mygreen}{RGB}{80,100,40}
\definecolor{myblue}{RGB}{30,90,100}

\definecolor{mygreen2}{RGB}{80,100,40}   

\usepackage{amsthm}
\theoremstyle{definition}

\usepackage{tabulary}
\newcolumntype{I}{!{\vrule width 1pt}}
\newcolumntype{x}[1]{>{\centering\arraybackslash}p{#1pt}}
\newcolumntype{y}[1]{>{\raggedright\arraybackslash}p{#1pt}}
\newcolumntype{z}[1]{>{\raggedleft\arraybackslash}p{#1pt}}
\newcommand{\tablestyle}[2]{\setlength{\tabcolsep}{#1}\renewcommand{\arraystretch}{#2}\centering\footnotesize}
\usepackage{amsmath}

\usepackage[pagebackref,breaklinks,colorlinks]{hyperref}

\usepackage[capitalize]{cleveref}
\crefname{section}{Sec.}{Secs.}
\Crefname{section}{Section}{Sections}
\Crefname{table}{Table}{Tables}
\crefname{table}{Tab.}{Tabs.}

\def\cvprPaperID{3522} 
\def\confName{CVPR}
\def\confYear{2023}

\begin{document}

\title{
TransFlow: Transformer as Flow Learner
}


\author[1, 2]{Yawen Lu}
\author[3]{Qifan Wang}
\author[4]{Siqi Ma}
\author[5]{Tong Geng}
\author[1]{Yingjie Victor Chen}
\author[6]{Huaijin Chen}
\author[2*]{Dongfang Liu}
\affil[1]{Purdue University}
\affil[2]{Rochester Institute of Technology}
\affil[3]{Meta AI}
\affil[4]{Westlake University}
\affil[5]{University of Rochester}
\affil[6]{Vayu Robotics}




\maketitle

\begin{abstract}
Optical flow is an indispensable building block for various important computer vision tasks, including motion estimation, object tracking, and disparity measurement. 
In\,this\,work, we propose TransFlow, a pure transformer architecture for optical flow estimation. Compared to dominant CNN-based methods, TransFlow demonstrates three advantages. First, it provides more accurate correlation and trustworthy matching in flow estimation by utilizing spatial self-attention and cross-attention mechanisms between adjacent frames to effectively capture global dependencies; Second, it recovers more compromised information ($e.g.,$ occlusion and motion blur) in flow estimation through long-range temporal association in dynamic scenes; Third, it enables a concise self-learning paradigm and effectively eliminate the complex and laborious multi-stage pre-training procedures. 
We achieve the state-of-the-art results on the Sintel, KITTI-15, as well as several downstream tasks, including video object detection, interpolation and stabilization.
For its efficacy, we hope TransFlow could serve as a flexible baseline for optical flow estimation.
\end{abstract}


\begin{figure}[t]
\centering
\includegraphics[width=1\columnwidth]{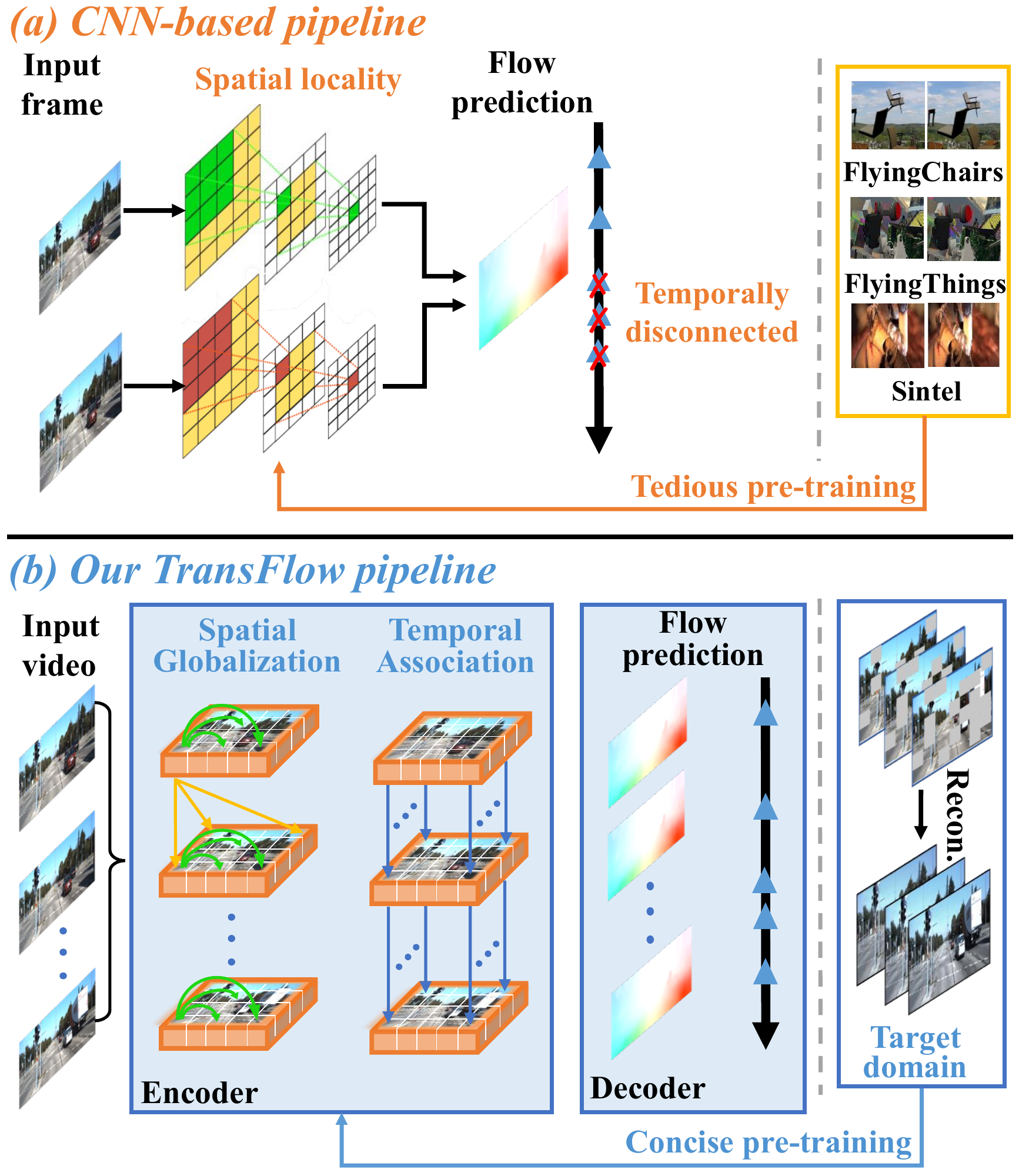}
\vspace{-5mm}
\caption{\textbf{Conceptual comparison of flow estimation methods.} Existing CNN-based methods regress flow via local spatial convolutions, while TransFlow relies on \textit{Transformer} to perform \textit{global matching} (both \textit{spatial} and \textit{temporal}). The demo video can be found in \url{https://youtu.be/xbnyj9wspqA}.} 
\vspace{-5mm}
\label{intro_compare}
\end{figure}

\vspace{-5mm}
\section{Introduction}
\label{sec:intro}
\vspace{-1mm}
With the renaissance of connectionism, rapid progress has been made in optical flow. Till now, most of the state-of-the-art flow learners were built upon Convolutional Neural Networks (CNNs) \cite{cao2022towards,gehrig2021raft,luo2021upflow, sun2018pwc, wang2020displacement,cheng2022physical}. Despite their diversified model designs and tantalizing results, existing CNN-based methods commonly rely on spatial locality and compute displacements from all-pair correlation volume for flow prediction (Fig.~\ref{intro_compare}(a)). Very recently, the vast success of Transformer \cite{vaswani2017attention,wang2022learning,wang2022visual} stimulates the emergence of attention-based paradigms for various tasks in language, vision, speech, and more. It appears to form a unanimous endeavor in the deep learning community to develop unified methodologies for solving problems in different areas.
Towards unifying methodologies, less inductive biases \cite{vaswani2017attention} are introduced for a specific problem, which urges the models to learn useful knowledge purely from the input data.


Jumping on the bandwagon of unifying architecture, we study applying
Vision Transformer \cite{dosovitskiy2020image} to the task of optical flow. 
The following question naturally arises: \textit{What are the major limitations of existing CNN-based approaches?} Tackling this question can provide insights into the model design of optical flow, and motivate us to rethink the task from an attention-driven view.\,First, the concurrent CNN-based methods demonstrate inefficiency in modeling \textit{global spatial dependencies} due to the intrinsic locality of the convolution operation. It usually requires a large number of CNN layers to capture the correlations between two pixels that are spatially far away.
Second, CNN-based flow learners generally model the flow between only two consecutive frames, and fail to explore \textit{temporal associations} in the neighboring contexts, resulting in weak prediction in the presence of significant photometric and geometric changes.\,Third, the existing training strategy usually requires a \textit{tedious pipeline}.\,Performance guarantees heavily rely on excessive pre-training on extra datasets (\eg,  FlyingChairs \cite{dosovitskiy2015flownet}, FlyingThings \cite{mayer2016large}, etc). Without adequate pre-training procedures, the model typically converges with large errors. 

In order to craft a Transformer architecture for optical flow that pursues performance guarantees, the question becomes more fundamental: \textit{How to address these limitations using Transformer?} 
As responses to this question, we articulate the technical contributions to address each of the above limitations:
 \vspace{-1mm}
 \begin{itemize}
    \item 
We introduce \textit{spatial attention} mechanism that effectively captures global dependencies and achieves precise correlation and reliable matching for flow estimation. 
Essentially, the spatial attention in Transformer enables effective contextual cue propagation from coherent regions to the surroundings with heavy-tailed noise, motion blurs, and large displacements, which significantly prevents performance degradation in flow estimation.
      \vspace{-5mm}
     \item  We explicitly model \textit{temporal association} in dynamic scenes using multi-frame features extracted in the designed Transformer encoder.
The correspondences among different frames are learned to generate the final estimated flow.
One advantage is that when a region in a frame is occluded or blurred, neighboring frames can effectively recover the missing information based on the learned temporal association.
 \vspace{-2mm}
    \item 
We
design a concise \textit{self-supervised pre-training} module that effectively eliminates the complex and laborious multi-stage pre-training procedures.
In particular, extended from  MAE \cite{he2022masked}, we develop a masking strategy during the training to adaptively mask out visual tokens and learn strong pixel representations by reconstructing clean signals from corrupted inputs.
 We demonstrate that the simple architecture results in a powerful entrance model, achieving stronger performance compared with SOTA baselines \cite{huang2022flowformer,teed2020raft, zhao2020maskflownet, luo2021upflow, jiang2021learning}. 

 \end{itemize} 
 
  \vspace{-2mm}
In summary, we re-formalize typical optical flow estimation within \textit{a pure transformer architecture --- TransFlow} (Fig.~\ref{intro_compare}(b)), which factorises pixel-wise flow learners with spatial dependencies and temporal associations to increase performance guarantees.
Remaining of the work is organized as follows: §\ref{sec:LR} provides literature review on the concurrent flow estimation methods. §\ref{sec:flowFormer} describes the model architecture of TransFlow. In §\ref{sec:Exp}, we detail the configuration setup and experimental settings. Concretely, §\ref{subsec:Sota}, shows that our method achieves impressive results in popular flow estimation datasets (\eg Sintel \cite{butler2012naturalistic} and KITTI 2015 \cite{menze2015joint}) and outperforms recent leading approaches; In §\ref{subsec:Ablations}, with a set of diagnostic experiment, our extensive experimental settings verify the effectiveness of our method. In §\ref{subsec:Down}, we demonstrate the transferability and generalizability of our method for modeling object motion for downstream tasks (\ie, video object detection, video interpolation, and video stabilization), which can benefit from our method without bells and whistles. 
In the end, we make conclusions in §\ref{sec:Concl} to highlight that this work is expected to pave the way for future research in this area.

\begin{figure*}[htb]
\centering
\includegraphics[width=1\textwidth]{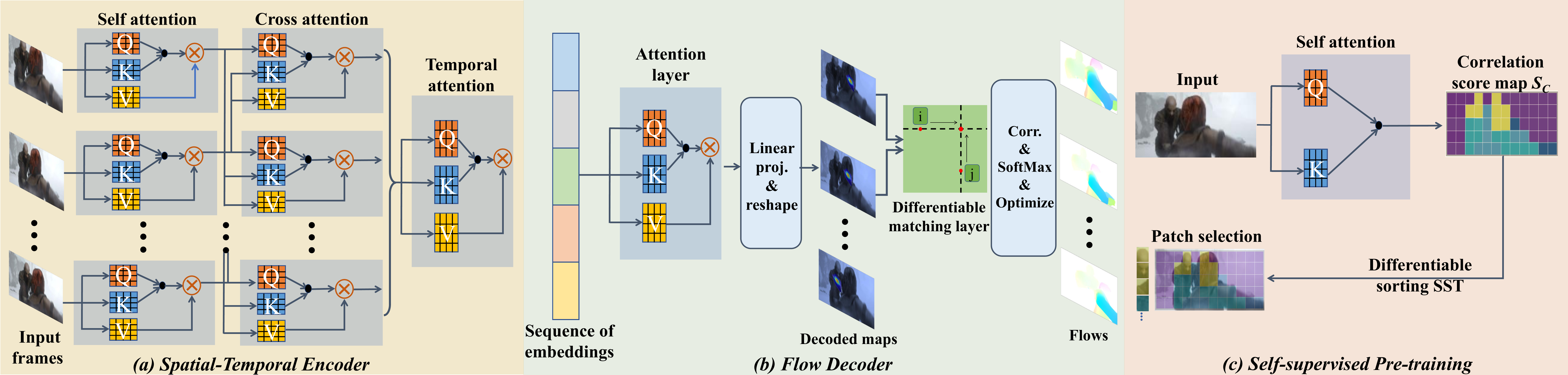}
\vspace{-3mm}
\caption{\textbf{Overall model architecture of TransFlow.} It consists of three major
components, a spatial-temporal encoder, a flow decoder,
and a self-supervised pre-training module. The spatial-temporal encoder jointly performs spatial globalization
and temporal association among patch tokens. The flow decoder decodes the feature maps
for multiple frames and generates the final optical flow. The pre-training module is designed to learn effective image representation
in a self-supervised manner.}
\vspace{-2mm}
\label{overview_framework}
\end{figure*}

\section{Important Knowns and Gap}
\vspace{-0.5mm}
\label{sec:LR}

\noindent \textbf{Optical Flow Learners.
}
Traditionally, optical flow was formulated as an energy optimization problem for maximizing similarity between image pairs \cite{black1993framework, bruhn2005lucas, sun2014quantitative}. More recently, visual similarity is computed via the computationally expensive correlation of high-dimensional features encoded by convolutional neural networks \cite{dosovitskiy2015flownet, ilg2017flownet, sun2018pwc, hofinger2020improving, teed2020raft, huang2022flowformer}. 
FlowNet \cite{dosovitskiy2015flownet} was the first end-to-end CNN-based network, which uses a coarse and refined branch for optical flow estimation. Its successive work, FlowNet2.0 \cite{ilg2017flownet}, adopted a stacked
architecture with warping operation, leading to improved performance. 
Following the
coarse-to-fine strategy,
PWC-Net \cite{sun2018pwc} developed a framework composed of stacked image pyramids, image warping and cost volumes; Hofinger et al., \cite{hofinger2020improving} replaced the image warping with a sampling-based strategy to improve the cost volume construction; Teed and Deng et al., \cite{teed2020raft} proposed to build a 4D cost volume for matching between all pairs of pixels and added a recurrent decoder for propagation. 
However, feature maps generated in these methods are usually suffering from a limited receptive field and high susceptibility to outliers, making them unsuitable as effective structures for learning global motion clues. Essentially, our method is conceptually different from these pioneer arts.
We design a transformer structure that takes advantage of both self- and cross- attentions and temporal association for effective global matching. \,Moreover, we\,demonstrate\,it is possible to achieve competitive results without costly training pipelines, using self-supervised learning.

\noindent \textbf{Attention in Optical Flow.}
While becoming the standard for natural language processing tasks \cite{devlin2018bert, dai2019transformer}, a flurry of research has successfully introduced Transformers to computer vision. Inspired by successes in image classification \cite{dosovitskiy2020image, yuan2021tokens}, multiple recent architectures have been trying to combine CNN-based architectures with self-attention, including detection \cite{carion2020end,cui2021tf}, image restoration \cite{wang2022uformer}, video inpainting \cite{liu2021fuseformer} and flow estimation \cite{sui2022craft}. Recently, there have been several attempts to apply Transformer structures to boost the performance of optical flow. Generally for these works, attention is applied in tandem with CNNs to compensate for the absence of image-specific inductive bias \cite{liu2021fuseformer, yang2022lite, wu2019pay, wu2021cvt}. A stack of Transformer blocks are added between CNN encoder and decoder for preventing blurry edges \cite{liu2021fuseformer} and a combination of light-weight self-attention and convolutions are unitized to improve the inconsistent segmentation output \cite{yang2022lite}. The most relevant work to ours is FlowFormer~\cite{huang2022flowformer}. It embeds the 4D cost volume into a latent feature with a transformer-based structure and decodes the latent feature with a convolutional recurrent network. 
There are some major differences between this work and ours. First, they only model two consecutive frames but ignore long-range temporal correlations. Second, we enable efficient and effective pre-training in optical flow, which fully explores the potential of the transformer model to rely on the target datasets only. More comparisons of these two are in the experiments.

\vspace{-1mm}
\section{TransFlow}
\label{sec:flowFormer}

The task of optical flow is to estimate a series of dense displacement fields from a sequence of consecutive frames. The overview of the TransFlow model architecture is illustrated in Fig.~\ref{overview_framework}. Our TransFlow is a transformer model that consists of three major components, a spatial-temporal encoder, a flow decoder and a self-supervised pre-training module.
The spatial-temporal encoder jointly performs spatial globalization and temporal association to effectively capture the correlations among frames and propagate global flow features. 
The flow decoder decodes the feature maps for multiple frames which are then used to generate the estimated optical flow.
The pre-training module is designed to learn the effective image pixel representation in a self-supervised manner, which eliminates the complex and laborious multi-stage pre-training procedures widely used in previous approaches.

\subsection{Problem Definition}
The input is a sequence of frames $X \in \mathbb{R}^{T \times H \times W \times C}$, which consists of $T$ frames and $C$ channels with $(H, W)$ as the resolution.
Following the work in ViT \cite{dosovitskiy2020image}, we split each frame into $N$ fixed-size non-overlapping patches $\mathbf{x}_{p}$, where $p\in \{1, 2, ..., N\}$, $h\times w$ is patch size, and $N = \frac{H}{h}\times \frac{W}{w}$ ensuring the $N$ patches span the entire frame. 
The purpose of TransFlow is to output a sequence of feature map $m$ for each frame, which is then used to generate the per-pixel displacement field $f$ between any source and target frames.

\subsection{Spatial-Temporal Encoder}

\paragraph{Spatial Globalization}
\label{subsec:spatial_att}
The existing CNN-based flow learners demonstrate inefficiency in modeling global spatial dependencies due to the intrinsic locality of the convolution operation. 
However, the global spatial correlation is important information which enables effective contextual cue propagation from coherent regions to the surroundings with heavy-tailed noise, motion blurs, and large displacements, preventing performance degradation in estimating the optical flow. 

In this work, we apply a spatial attention mechanism between two consecutive frames to capture the global spatial dependencies among the pixels.
In particular, similar to ViT \cite{dosovitskiy2020image}, each patch $\mathbf{x}_{p}$ (\cf Table~\ref{patch_size} for patch size) is first converted into a $d$-dimensional embedding vector $\mathbf{e}_{p}\in \mathbb{R}^d$ with a projection matrix $\mathbf{W}_e$. 
The final input sequence of patch embeddings is denoted as:
\begin{equation}
\begin{aligned}
\mathbf{z}^0 &=\left[\mathbf{e}_{1}, \mathbf{e}_{2}, \dots, \mathbf{e}_{N}\right];\\
\mathbf{e}_{p} &= \mathbf{W}_e \cdot \mathbf{x}_{p} + \mathbf{p}_{p},
\end{aligned}
\end{equation}
where $\mathbf{p}_{p}$ is a set of learnable position embeddings (\cf Table~\ref{positional_embed}) to retain the positional information, which is significant to motion clues.
The patch tokens are passed through a series of Multi-head Self-Attention (MSA), Multi-head Cross-Attention (MCA) and MLP layers:
\begin{equation}
\label{spatial_embedding}
\begin{aligned}
\mathbf{y}^{\ell} &=\operatorname{MLP}\left(\operatorname{MSA}\left(\mathbf{z}^{\ell-1}\right)\right)+\mathbf{z}^{\ell-1}; \\
\mathbf{z}^{\ell} &=\operatorname{MLP}\left(\operatorname{MCA}\left(\mathbf{y}^{\ell}\right)\right)+\mathbf{y}^{\ell}.
\end{aligned}
\end{equation}
Essentially, the self-attention is used to capture the global pixel dependencies within the same frame, while the cross-attention is designed to communicate the information between two adjacent frames. The self-attention and cross-attention are defined as:
\begin{equation}\label{eq:self-attention}
\begin{aligned}
&\operatorname{MSA}\left(\mathbf{z}\right) = softmax\left({\mathbf{Q}\cdot \mathbf{K}^T}/{\sqrt{d}}\right)\cdot{\mathbf{V}};\\
&\operatorname{MCA}\left(\mathbf{z}\right) = softmax\left({\mathbf{Q}\cdot {\mathbf{K}^{\prime}}^T}/{\sqrt{d}}\right)\cdot{\mathbf{V}^{\prime}},
\end{aligned}
\end{equation}
where $\mathbf{Q}=\mathbf{z}\cdot\mathbf{W}^Q$, $\mathbf{K}=\mathbf{z}\cdot\mathbf{W}^K$ and $\mathbf{V}=\mathbf{z}\cdot\mathbf{W}^V$ are the query, key and value embedding matrices in MSA. $\mathbf{K}^{\prime}=\mathbf{z}^{\prime}\cdot{\mathbf{W}^{\prime}}^K$ and $\mathbf{V}^{\prime}=\mathbf{z}^{\prime}\cdot{\mathbf{W}^{\prime}}^V$ are the key and value embedding matrices in MCA. 

The output from the attention layers is the refined correlation features, and we interleave the self-attention and cross-attention layers by $L$ times. Through the joint aggregation, we benefit from the feature aggregation via local frame in self-attention, which is further facilitated via adjacent perspectives in cross-attention, as depicted in Fig.~\ref{overview_framework} (a).




\vspace{-2mm}
\paragraph{Temporal Association}
\label{subsec:temp_att}
\vspace{-1mm}
Previous approaches for estimating optical flow from each pair of adjacent frames are less effective as they ignore the inherent nature of long-range temporal associations. The motion estimation in discontinuous and occluded regions cannot be well modelled under modern architectures. To better capture high-level temporal information in the flow tokens, we learn the token embeddings by jointly modeling the temporal association with the spatial attention described above. As a result, each transformer layer can measure long-range interaction between input embeddings. Specifically, given a sequence of attentioned features from video clips consisting of $T$ frames (see related experiments in Table \ref{temporal_length}), we iteratively choose one as query and the rest as key features to compute the temporal attention using Softmax, which is similar as Eq. \ref{eq:self-attention}. Resulting in a $d$-dimensional embedded feature volume $\textbf{z} \in \mathbb{R}^{T \times h \times w \times d}$, the feature volume is then passed to the following transformer decoding block. By learning temporal festures in this way, temporal information can be accumulated into each frame to capture temporal associations across frames, which is shown in Fig.~\ref{overview_framework} (a).

\subsection{Flow Decoder}
\label{flow_generation}
Different from the traditional Transformer decoder, our decoder is designed to decode the feature maps of all frames (Fig.~\ref{overview_framework} (b)). These decoded features are then utilized in obtaining the final flows.
Therefore, our decoder aims to generate multiple feature maps at the same time with a fixed sequence length, instead of autoregressive decoding. There are two major advantages in such a design. First, simultaneous decoding allows us to remove the encoder-decoder cross-attention in the traditional Transformer decoders. Second, beam search is no longer needed, which makes our decoding process much more efficient. Therefore, in this work, we adopt a structurally symmetric design with the Transformer encoder. In other words, our decoder has the same self-attention architecture as our encoder except that the input to the decoder is the latent cost embedding from the encoder. 

Given the decoded feature maps between two consecutive frames, we compare the feature similarity by computing the correlation following \cite{wang2020learning}. To enable the end-to-end training, we apply the differentiable matching layer \cite{xu2020aanet} to identify the correspondence from the adjacent frames. The final flow $f$ can then be generated from the correspondences. During training, we further conduct an additional occlusion detection \cite{feng2022disentangling} by performing a forward consistency checking and considering pixels to be occluded if the mismatching in both frames is too large.
Consequently, the occlusion areas $M_{occ}$ is computed as $M_{occ} = f_{D}(I_{s} - I_{t}(x+f))$, where $f_{D}$ can be any function that measures the photometric distance. $f$ is the estimated forward optical flow. $I_s$ and $I_t$ are the source and target images/frames. The overall objective can be formulated as:
\begin{small} 
\begin{equation}
\label{occlusion_loss}
\begin{aligned}
L =\sum_{i=1}^R (1-M_{occ}) \gamma^{(i-R)}\left\|f_{gt}-f\right\|_1,
\end{aligned}
\end{equation}
\end{small}where $R$ is the total number of the training iterations. $\gamma$ is a hyperparameter that controls the weight of the loss among different iterations. $f_{gt}$ stands for the provided ground-truth flow map.

\subsection{Self-supervised Pre-training}
\label{subsec:strategical}
The performance of the existing flow learners heavily relies on excessive pre-training on extra synthetic datasets, followed by fine-tuning on the target domain. Without adequate pre-training procedures on large-scale data, the model typically converges with large errors. 
Therefore, it is an important task to design an efficient and effective pre-training strategy that improves the downstream optical flow task.

Inspired by the recent MAE \cite{he2022masked}, we introduce a masking strategy in self-supervised pre-training that adaptively masks out patch tokens and learns pixel representations by reconstructing clean signals from corrupted inputs.
Specifically, we learn a score map for patch selection to choose the most informative patches as masked tokens under a determined ratio, as opposed to randomly masking in \cite{he2022masked} or uniformly masking in \cite{li2022uniform}. In our diagnostic experiments (\cf Table~\ref{sampling_strategy} and \ref{masking_ratio}), we will demonstrate that the capability of our self-learning paradigm in recovering crucial regions can be enhanced. More specifically, we adopt multiple layers of self-attention blocks taking all the patch token embeddings as input. 
The attention map is then calculated as the correlation between the query embedding from the image token $Q$ and all key embeddings across all patches $K$. The correlations are then followed by a Softmax activation to generate the correlation score map $S_{c}$, as depicted in Eq. \ref{strategical_masking}. The correlation score map output from the final layer of the attention blocks will be utilized to guide our strategic masking learning:
\begin{equation}
\label{strategical_masking}
S_{c}= softmax (Q \cdot {K^{T}})
\end{equation}






The obtained correlation score map $S_{c}$ is then modeled as a ranking problem to be sorted in an ascending order to select the most informative tokens for masking, as in Fig.~\ref{overview_framework} (c). In order to prevent the discrete property of the $argsort$ operation, we instead utilize the soft sort operation in \cite{prillo2020softsort} denoting as $SST(\cdot)$:
\begin{small} 
\begin{equation}
\label{strategical_masking_sort}
\begin{aligned}
SST(\cdot) = softmax(\frac{|sort(S_{c}) \textbf{1}^{T}-\textbf{1}(S_{c})^{T}|}{\tau})
\end{aligned}
\end{equation}
\end{small}

\noindent where $|\cdot|$ calculates element-wise absolute and $\tau$ is the temperature constant that is set to 0.1 to control the degree of approximation. With the differentiable sorting, we are able to identify and retain the most significant token candidates and adaptively learn the score map as network weights in conjunction with our primary task.

\begin{table*}[t]
\centering
\tablestyle{8pt}{0.96}
\scalebox{0.96}{
\begin{tabular}{c|r|cc|cc|cc|cl}
\hline
\rowcolor{mygray}
\cellcolor{mygray}                              & \cellcolor{mygray}                      & \multicolumn{2}{c|}{\cellcolor{mygray}Sintel (train)} & \multicolumn{2}{c|}{\cellcolor{mygray}KITTI-15 (train)} & \multicolumn{2}{c|}{\cellcolor{mygray}Sintel (test)} & \multicolumn{2}{c}{\cellcolor{mygray}KITTI-15 (test)} \\ \cline{3-10} 
\rowcolor{mygray}
\multirow{-2}{*}{\cellcolor{mygray}Training Data} & \multirow{-2}{*}{\cellcolor{mygray}Method} & clean           & final          & F1-epe           & F1-all          & clean           & final           & \multicolumn{2}{c}{\cellcolor{mygray}F1-all}        \\ \hline \hline
\multirow{12}{*}{C+T}                                                    & PWC-Net \textcolor{lightgray}{\scriptsize{[CVPR18]}} \cite{sun2018pwc}        & 2.55            & 3.93           & 10.35            & 33.7            & -               & -               & \multicolumn{2}{c}{-}               \\
                                                                         & HD3 \textcolor{lightgray}{\scriptsize{[CVPR19]}} \cite{yin2019hierarchical}             & 3.84            & 8.77           & 13.17            & 24.0            & -               & -               & \multicolumn{2}{c}{-}              \\
                                                                         & LiteFlowNet \textcolor{lightgray}{\scriptsize{[TPAMI20]}} \cite{hui2018liteflownet}   & 2.24            & 3.78           & 8.97             & 25.9            & -               & -               & \multicolumn{2}{c}{-}               \\
                                                                         & RAFT \textcolor{lightgray}{\scriptsize{[ECCV20]}} \cite{teed2020raft}            & 1.43            & 2.71           & 5.04             & 17.4            & -               & -               & \multicolumn{2}{c}{-}               \\
                                                                         & FM-RAFT \textcolor{lightgray}{\scriptsize{[ECCV21]}} \cite{jiang2021learningFEW}       & 1.29            & 2.95           & 6.80             & 19.3            & -               & -               & \multicolumn{2}{c}{-}               \\
                                                                         & GMA \textcolor{lightgray}{\scriptsize{[ICCV21]}} \cite{jiang2021learningGMA}            & 1.30            & 2.74           & 4.69             & 17.1            & -               & -               & \multicolumn{2}{c}{-}               \\
                                                                         & Separable Flow \textcolor{lightgray}{\scriptsize{[ICCV21]}} \cite{zhang2021separable} & 1.30            & 2.59           & 4.60             & 15.9            & -               & -               & \multicolumn{2}{c}{-}               \\
                                                                         & Flow1D \textcolor{lightgray}{\scriptsize{[ICCV21]}} \cite{xu2021high}         & 1.98            & 3.27           & 5.59             & 22.95           & -               & -               & \multicolumn{2}{c}{-}              \\
                                                                         & AGFlow \textcolor{lightgray}{\scriptsize{[AAAI22]}} \cite{luo2022learningAGflow}          & 1.31            & 2.69           & 4.82             & 17.0            & -               & -               & \multicolumn{2}{c}{-}              \\
                                                                         & KPA-Flow \textcolor{lightgray}{\scriptsize{[CVPR22]}} \cite{luo2022learningKPA}       & 1.28            & 2.68           & 4.46             & 15.9            & -               & -               & \multicolumn{2}{c}{-}              \\
                                                                         & Flowformer \textcolor{lightgray}{\scriptsize{[ECCV22]}} \cite{huang2022flowformer}     & \underline{1.01}            & \underline{2.40}           & \underline{4.09}             & \underline{14.72}           & -               & -               & \multicolumn{2}{c}{-}               \\
                                                                         & \textbf{Ours}                                                                                 & \textbf{0.93}   & \textbf{2.33}  & \textbf{3.98}    & \textbf{14.4}   & -               & -               & \multicolumn{2}{c}{-}               \\ \hline
\multirow{10}{*}{\begin{tabular}[c]{@{}c@{}}C+T+S+K\\ (+H)\end{tabular}} & PWC-Net \textcolor{lightgray}{\scriptsize{[CVPR18]}} \cite{sun2018pwc}        & -               & -              & -                & -               & 4.39            & 5.04            & \multicolumn{2}{c}{9.60}          \\
                                                                         & HD3 \textcolor{lightgray}{\scriptsize{[CVPR19]}} \cite{yin2019hierarchical}           & 1.87            & 1.17           & 1.31             & 4.1             & 4.79            & 4.67            & \multicolumn{2}{c}{6.55}          \\
                                                                         & LiteFlowNet \textcolor{lightgray}{\scriptsize{[TPAMI20]}} \cite{hui2018liteflownet}  & 1.35            & 1.78           & 1.62             & 5.58            & 4.54            & 5.38            & \multicolumn{2}{c}{9.38}          \\
                                                                         & RAFT \textcolor{lightgray}{\scriptsize{[ECCV20]}} \cite{teed2020raft}            & 0.77            & 1.20           & 0.64             & 1.5             & 2.08            & 3.41            & \multicolumn{2}{c}{5.27}          \\
                                                                         & FM-RAFT \textcolor{lightgray}{\scriptsize{[ECCV21]}} \cite{jiang2021learningFEW}        & 0.86            & 1.75           & 0.75             & 2.1             & 1.77            & 3.88            & \multicolumn{2}{c}{6.17}          \\
                                                                         & Separable Flow \textcolor{lightgray}{\scriptsize{[ICCV21]}} \cite{zhang2021separable} & 0.71            & 1.14           & 0.68             & 1.57            & 1.99            & 3.27            & \multicolumn{2}{c}{4.89}          \\
                                                                         & Flow1D \textcolor{lightgray}{\scriptsize{[ICCV21]}} \cite{xu2021high}        & (0.84)          & (1.25)         & -                & (1.6)           & (2.24)          & (3.81)          & \multicolumn{2}{c}{(6.27)}        \\
                                                                         & KPA-Flow \textcolor{lightgray}{\scriptsize{[CVPR22]}} \cite{luo2022learningKPA}      & (0.60)          & (1.02)         & (\underline{0.52})           & (\underline{1.10})          & (1.35)          & (2.36)          & \multicolumn{2}{c}{(\underline{4.60})}        \\
                                                                         & Flowformer \textcolor{lightgray}{\scriptsize{[ECCV22]}} \cite{huang2022flowformer}    & (\underline{0.48})          &(\underline{0.74})         & (0.53)           & (1.11)          & (\underline{1.16})          & (\underline{2.09})          & \multicolumn{2}{c}{(4.68)}        \\
                                                                         & \textbf{Ours}                                                                                 & (\textbf{0.42})   & (\textbf{0.69})  & (\textbf{0.49})    & (\textbf{1.05})   & (\textbf{1.06})   & (\textbf{2.08})   & \multicolumn{2}{c}{(\textbf{4.32})} \\ \hline
\end{tabular}}
\vspace{-2mm}
\captionsetup{font=small}
\caption{\textbf{Quantitative comparisons with state-of-the-arts.} We follow existing works to compare the results on two standard benchmarks Sintel and KITTI-15. "C+T" denotes training only on FlyingChairs and FlyingThings datasets and testing on others for the generalization ability. "C+T+S+K(+H)" denotes training on mixed datasets and testing on Sintel and KITTI-15 for evaluation. Recent works \cite{xu2021high, luo2022learningKPA, huang2022flowformer} including HD1K \cite{kondermann2016hci} dataset for training are marked with brackets in results. Our self-learning paradigm helps to get superior results by avoiding tedious pre-training stages on "C/T" and simplifying the training pipeline. The best and second best results are highlighted in bold and underlined. See \S \ref{subsec:Sota} for details.}
\label{big_table}
\vspace{-2.5mm}
\end{table*}

\vspace{-1mm}
\section{Experiments}
\label{sec:Exp}
\vspace{-1mm}
\noindent \textbf{Datasets.}
\label{subsec:Dataset}
Existing flow estimation approaches require a tedious training pipeline which first pre-train the models on FlyingChairs (“C”) \cite{dosovitskiy2015flownet} and FlyingThings (“T”) \cite{mayer2016large}, and then fine-tune the trained models on Sintel (“S”) \cite{butler2012naturalistic} and KITTI 2015 (“K”) \cite{menze2015joint}. Without the progressive steps, the flow estimation performance will get a significant degradation. Simplifying the 
cumbersome procedures, we rely on training optical flow task on the target domain without excessive pre-training stages. 
MPI-Sintel \cite{butler2012naturalistic} dataset is rendered based on animated movies and is split into \textit{Clean} and \textit{Final} pass. KITTI-15 \cite{menze2015joint}  contains 200 training and 200 testing road scenes with sparse ground truth flow, where images are captured via stereo cameras. For datasets provide only pairwise flow (e.g., KITTI-15), we access raw data in the self-supervised pre-training. 


\vspace{0.5mm}
\noindent \textbf{Implementation Details.}
\label{subsec:Setup}
We stack 12 transformer blocks in the encoder to adaptively learn the feature encoding. To keep the resolution to be the same as the input, we adopt the convex upsampling technique in \cite{teed2020raft} to upsample the prediction. The model is first pre-trained in a self-learning paradigm with a learning rate of 1$e$-4 and then the entire network is continuously trained on the target domain with a batch size of 6 and learning rate of 12.5$e$-5 for 140K steps. For the hyperparameters, $\gamma$ is set to 0.8 and the masking ratio is 50\%.
The detailed diagnostic experiments of these hyperparameters are provided in §\ref{subsec:Ablations}.

\vspace{0.5mm}
\noindent \textbf{Evaluation Metrics.}
\label{subsec:Metrics}
The main evaluation metrics, used by the Sintel datasets, is the average end-point error (\textit{AEPE}), which denotes the average pixel-wise flow error. The KITTI dataset adopts \textit{F1-epe} (\%) and \textit{F1-all} (\%), which refers to the percentage of flow outliers over all pixels on foreground regions and entire image pixels.

\subsection{Comparison to the State-of-the-Art Methods}
\label{subsec:Sota}

\noindent \textbf{Quantitative Evaluations.} We compare our approach with existing supervised flow estimation methods on the most popular optical flow benchmarks (\ie Sintel and KITTI). Without tedious multi-stage flow estimation pre-training on synthetic benchmarks FlyingChairs and FlyingThings, our designated framework beats existing state-of-the-art methods, as demonstrated by the quantitative results in Table~\ref{big_table}. As shown, for generalization ability, we train our TransFlow on the FlyingChairs and FlyingThings (C+T) and directly evaluate it on the Sintel and KITTI-15 without further fine-tuning. Our TransFlow depicts the best result with the smallest errors among all compared methods on both datasets. Specifically on the Sintel dataset, we achieves \textbf{0.93} and \textbf{2.33} \textit{AEPE} on the clean and final pass, which is \textbf{0.50} and \textbf{0.38} lower than the widely used method RAFT \cite{teed2020raft}. On the KITTI-15 dataset, we reduce the \textit{F1-all} error by \textbf{17.2\%} of RAFT \cite{teed2020raft}. 

When following the introduced self-learning paradigm on the target datasets and evaluate on the Sintel test set, our method achieves a \textbf{1.06} and \textbf{2.08} \textit{AEPE} on the Sintel clean and final pass, which is \textbf{49\%} and \textbf{39\%} lower than RAFT \cite{teed2020raft}, respectively. Similarly on the KITTI-15 benchmark, our approach performs a \textbf{4.32} \textit{F1-all} score in errors, which is \textbf{0.95} and \textbf{0.36} lower than recent RAFT \cite{teed2020raft} and FlowFormer \cite{huang2022flowformer}. However, RAFT \cite{teed2020raft} and FlowFormer \cite{huang2022flowformer} both require a multi-stage flow estimation pre-trainings before training on C+T+S+K or C+T+S+K+H datasets. Compared to those existing approaches, our proposed pipeline delivers superior performance with significantly simpler and more effective steps.




\begin{figure*}[tb]
\centering
\includegraphics[width=0.89\textwidth]{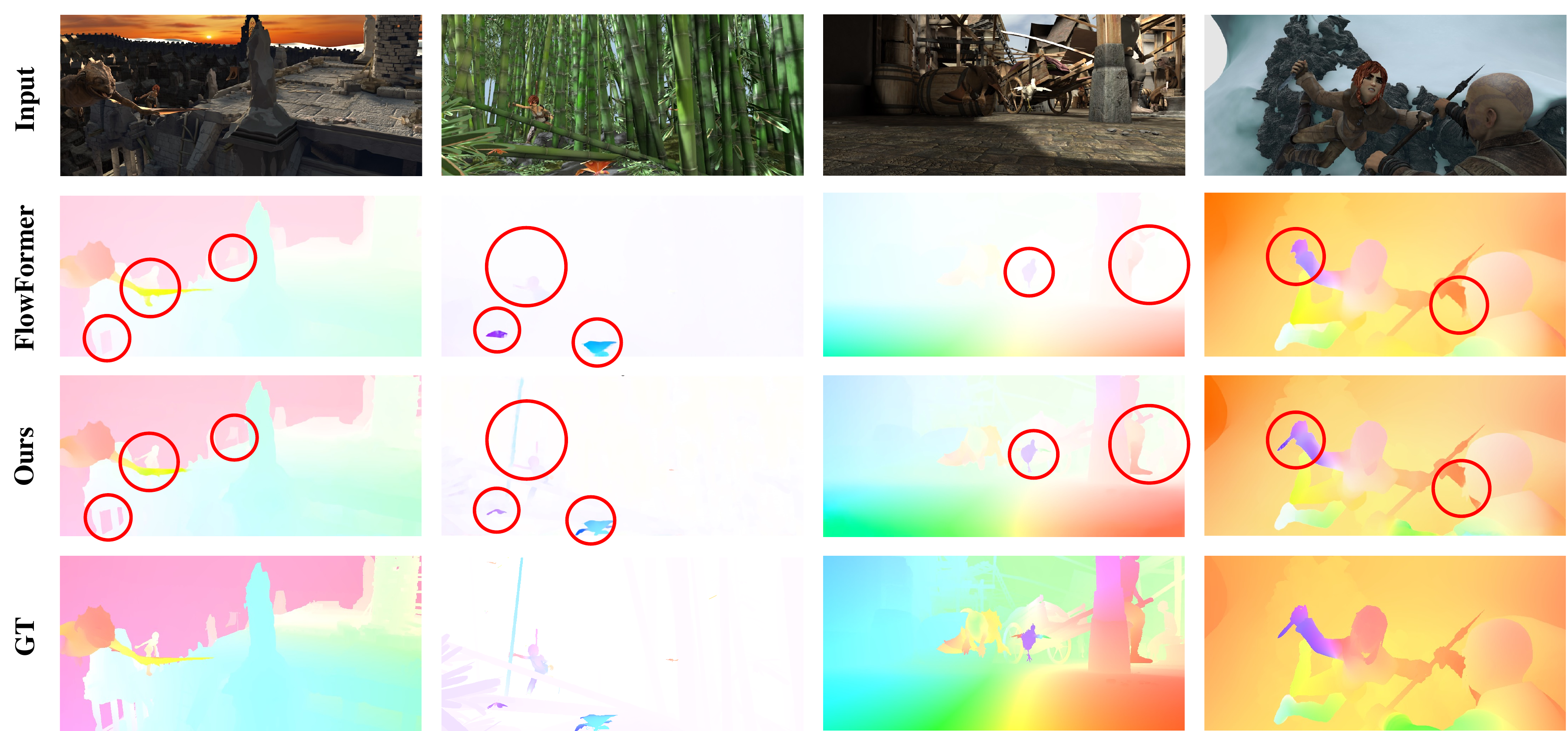}
\vspace{-0.5mm}
\caption{\textbf{Qualitative results on Sintel \texttt{val} set.} Given the target frame, we show the results of the state-of-the-art FlowFormer \cite{huang2022flowformer}, our TransFlow results, and the provided ground truth flow. \textbf{\tikzcircle[red, radius=1.5cm]{3pt}} highlights comparing details. See \S \ref{subsec:Sota} for details.}
\vspace{-2mm}
\label{qualitative_fig}
\end{figure*}

\vspace{0.5mm}
\noindent \textbf{Qualitative Evaluations.} We sampled test samples from Sintel \texttt{val} set and provide the corresponding optical flow estimation of the state-of-the-art FlowFormer \cite{huang2022flowformer} and our TransFlow in Fig.~\ref{qualitative_fig}. As shown, our TransFlow shows superior capability in distinguishing occluded regions and performing clearer boundaries on small and thin objects, benefiting from our consideration in the spatial globalization and temporal associations. These enhancements demonstrate the efficacy of our designed structures in spatial attentions, temporal associations, and novel strategic masking strategy in improving flow reasoning. 

\subsection{Diagnostic Experiments}
\label{subsec:Ablations}

\noindent \textbf{Core Components.} First, we study the efficacy of the core components of our algorithm in Table~\ref{co-component} to analyze their contributions to the final results. As shown, It can be seen that solely self-attention can yield limited performance as a baseline of our framework, while a combination of both self- and cross- attentions boost the performance thanks to the effective local feature aggregation between two views. The design of temporal association among multiple frames successfully alleviates the potential ambiguities in un-smooth and occluded regions, therefore bringing further improvement. The strategic masking reduces the reliance on multi-stage pre-training and benefits our self-learning paradigm. The occlusion consistency on the flow loss has a similar effect with the temporal association and adds additional performance gains to the results. From Table~\ref{co-component}, each component contributes to the improvement of performance, clearly depicting the effectiveness of our proposed components.

\noindent \textbf{Patch Settings.} We empirically evaluate the end-point error by
adjusting the patch size (\cf §\ref{subsec:spatial_att}) in our TransFlow. As shown in Table~\ref{patch_size}, when patch size are increasingly set from 4 $\times$ 4 to 8 $\times$ 8, the performance are slightly improved (\textbf{0.44} and \textbf{0.71} $\rightarrow$ \textbf{0.42} and \textbf{0.69}) in \textit{EPE} and \textbf{1.11} $\rightarrow$ \textbf{1.05} in \textit{F1-all} on Sintel and KITTI datasets, respectively. Nevertheless, when further enlarging the patch size to 16 $\times$ 16, the performance drops while the cost of computing continues to drop. The reason is that larger patch sizes lead to bigger kernel regions, resulting in the loss of global and long-range context information which is crucial for flow propagation.

\begin{table*}[tb]
\vspace{-2.5mm}
\begin{subtable}{0.5\linewidth}
\centering
\caption{Contribution of each core component (\cf §\ref{sec:flowFormer}).}
\scalebox{0.78}{
\begin{tabular}{cccccccc}
\rowcolor{mygray}
\hline
\multicolumn{5}{c}{Config}                                                                                                                        & \multicolumn{2}{c}{Sintel (val)} & KITTI (val) \\ \hline
\rowcolor{mygray}
Self                      & Cross                       & Temporal                    & Masking                     & Occ                         & Clean           & Final          & F1-all      \\ \hline  \hline
\usym{2713} &  &  &  &  & 0.58            & 0.81           & 1.22        \\
\usym{2713} & \usym{2713}   &  &  &  & 0.55            & 0.78           & 1.18        \\
\usym{2713} & \usym{2713}   & \usym{2713}   &   &   & 0.49            & 0.73           & 1.18        \\
\usym{2713} & \usym{2713}   & \usym{2713}  & \usym{2713}   &   & 0.44            & 0.70           & 1.07        \\
\textcolor{red}{\usym{2713}} & \textcolor{red}{\usym{2713}}   & \textcolor{red}{\usym{2713}}   & \textcolor{red}{\usym{2713}}   & \textcolor{red}{\usym{2713}}   & \textbf{0.42}            & \textbf{0.69}           & \textbf{1.05}        \\ \hline
\end{tabular}
\label{co-component}} 
\end{subtable} 
\begin{subtable}{0.52\linewidth}  
\centering
\caption{Patch size settings (\cf §\ref{subsec:spatial_att}).}
\scalebox{0.785}{
\begin{tabular}{ccccc}
\hline
\rowcolor{mygray}
\cellcolor{mygray}                             & \multicolumn{2}{c}{\cellcolor{mygray}Sintel (val)} & \multicolumn{2}{c}{\cellcolor{mygray}KITTI (val)} \\ \cline{2-5} 
\rowcolor{mygray}
\multirow{-2}{*}{\cellcolor{mygray}Patch size} & Clean                       & Final                      & \multicolumn{2}{c}{\cellcolor{mygray}F1-all}      \\ \hline  \hline
4 $\times$ 4                       & 0.44             & 0.71            &\quad \ \ 1.11          \\
\textcolor{red}{8 $\times$ 8}                       & \textbf{0.42}             & \textbf{0.69}            &\quad \ \ \textbf{1.05}          \\
14 $\times$ 14                     & 0.46             & 0.77            &\quad \ \ 1.14          \\
16 $\times$ 16                     & 0.59             & 0.85            &\quad \ \ 1.21          \\ \hline
\end{tabular}
\label{patch_size}} 
\end{subtable}

\vspace{2.5mm}
\begin{subtable}{0.28\linewidth}  
\centering
\caption{Different Pos. embeddings (\cf §\ref{subsec:spatial_att}).}
\scalebox{0.6}{
\begin{tabular}{ccccc}
\hline
\rowcolor{mygray} 
\cellcolor{mygray}                                                                         & \multicolumn{2}{c}{\cellcolor{mygray}Sintel (val)} & \multicolumn{2}{c}{\cellcolor{mygray}KITTI (val)} \\ \cline{2-5} 
\rowcolor{mygray}
\multirow{-2}{*}{\cellcolor{mygray}Pos. Embed} & Clean                       & Final                      & \multicolumn{2}{c}{\cellcolor{mygray} F1-all}      \\ \hline \hline
Fixed Abs Pos.                                                            & 0.43             & 0.71            &\quad \ \ 1.06          \\
\textcolor{red}{Learnable Abs Pos.}                                                        & \textbf{0.42}             & \textbf{0.69}            &\quad \ \ \textbf{1.05}          \\
PEG Pos.                               & \textbf{0.42}             & 0.70            &\quad \ \ 1.07          \\
Learnable Rel Pos.                                                        & 0.47             & 0.76            &\quad \ \ 1.10          \\ \hline
\end{tabular}
\label{positional_embed}}
\end{subtable}
\vspace{1.5mm}
\begin{subtable}{0.22\linewidth}  
\centering
\caption{Temporal Length (\cf §\ref{subsec:temp_att}).}
\scalebox{0.6}{
\begin{tabular}{ccc}
\hline
\rowcolor{mygray}  
\cellcolor{mygray}                                  & \multicolumn{2}{c}{\cellcolor{mygray} Sintel (val)} \\ \cline{2-3} 
\rowcolor{mygray}  
\multirow{-2}{*}{\cellcolor{mygray}Temporal length}               & Clean            & Final           \\ \hline \hline
2 frame                          &     0.47             &   0.75              \\               
3 frame                          &     0.44             &   0.73              \\
\textcolor{red}{5 frame}                          &     \textbf{0.42}             &   0.69              \\
7 frame                        &     0.43             &   \textbf{0.68}              \\ \hline
\end{tabular}
\label{temporal_length}}
\end{subtable}
\vspace{1.5mm}
\begin{subtable}{0.245\linewidth}  
\centering
\caption{Sampling methods (\cf §\ref{subsec:strategical}).}
\scalebox{0.6}{
\begin{tabular}{ccccc}
\hline
\rowcolor{mygray}
\cellcolor{mygray}                             & \multicolumn{2}{c}{\cellcolor{mygray}Sintel (val)} & \multicolumn{2}{c}{\cellcolor{mygray}KITTI (val)} \\ \cline{2-5} 
\rowcolor{mygray}
\multirow{-2}{*}{\cellcolor{mygray}Sampling} & Clean                       & Final                      & \multicolumn{2}{c}{\cellcolor{mygray}F1-all}      \\ \hline  \hline
Block                    &     0.48             &     0.74            & \quad \ \  1.10    &             \\
Random                    &     0.50             &     0.76            & \quad \ \  1.13    &             \\
Uniform                   &      0.45            &     0.72            & \quad \ \ 1.09    &             \\
\textcolor{red}{Strategic}                      &    \textbf{0.42}              &      \textbf{0.69}           &   \quad \ \  \textbf{1.05}   &             \\ \hline
\end{tabular}
\label{sampling_strategy}}
\end{subtable}
\vspace{1.5mm}
\begin{subtable}{0.24\linewidth}  
\centering
\caption{Masking ratio (\cf §\ref{subsec:strategical}).}
\scalebox{0.6}{
\begin{tabular}{ccccc}
\hline
\rowcolor{mygray}
\cellcolor{mygray}                                                                            & \multicolumn{2}{c}{\cellcolor{mygray} Sintel (val)} & \multicolumn{2}{c}{\cellcolor{mygray}KITTI (val)} \\ \cline{2-5} 
\rowcolor{mygray}
\multirow{-2}{*}{\cellcolor{mygray}Masking ratio} & Clean                       & Final                      & \multicolumn{2}{c}{\cellcolor{mygray}F1-all}      \\ \hline \hline
30\%                                                                      &         0.45         &    0.71             &  \quad \ \      1.08       \\
\textcolor{red}{50\%}                                                                        &         \textbf{0.42}         &    \textbf{0.69}             & \quad \ \  \textbf{1.05}       \\
70\%                                                                    &         0.46         &    0.71             &    \quad \ \    1.09       \\
90\%                                                                      &          0.49        &      0.77           &  \quad \ \   1.16          \\ \hline
\end{tabular}
\label{masking_ratio}}
\end{subtable}
\vspace{-4mm}
\caption{\textbf{A set of diagnostic experiments.} The adopted algorithm designs and settings are marked in \textcolor{red}{red}. See \S \ref{subsec:Ablations} for details.}
\vspace{-2mm}
\end{table*}


\noindent \textbf{Positional Embedding.}
The efficacy of different positional embeddings (\cf §\ref{subsec:spatial_att}) is rarely discussed in previous works. Consequently, we compare the performance of flow estimation under different positional embeddings (\eg Abs/Rel, Learnable/Fixed). As depicted in Table~\ref{positional_embed}, we observe that the learnable Abs Pos. achieves a slightly better result than the fixed Abs Pos. and a recent Positional Encoding
Generator (PEG) \cite{chu2021conditional}, while showing a larger improvement than relative Pos. We suppose that the fixed sin-cos Abs Pos. can encode the flow features almost as well as the learnable Abs Pos. and PEG Pos., and positional embedding is indispensable in our setting. In addition, we believe the degradation from relative Pos. is due to the fact that object motion requires more absolute position encoding in order to locate and learn the motion, and global information is also more important than local relative information in this task, which is consistent with our claim.

\noindent \textbf{Temporal Length.}
Table~\ref{temporal_length} shows that as we increase the number of frames (\cf §\ref{subsec:temp_att}) fed into our temporal module, the error gets decrease since the network is able to incorporate longer temporal context and to avoid temporal artifacts and discontinuous estimation in the flow. However, Table~\ref{temporal_length} also demonstrates that the accuracy will become saturated once the number of temporal length is sufficient enough to cover visible motion. Considering that when increasing the temporal length from 5 to 7, there is no discernible difference in performance while the computational cost will increase correspondingly. Therefore, we choose 5-frame length as input.


\noindent \textbf{Sampling Strategy}.
Table~\ref{sampling_strategy} shows the effect of various sampling strategies for masking (\cf §\ref{subsec:strategical}). We compare our strategic masking with block-wise masking \cite{bao2021beit}, random masking \cite{he2022masked} and uniform masking \cite{li2022uniform}. Under the same masking ratio, it can be seen the compared samplings have different levels of degradation compared to ours. The naive block-wise masking and random masking may destroy the tokens of vital regions of the original image that are required for object motion, whereas uniform masking may disregard the significance and relationship between tokens. On the contrary, our sampling has the ability to learn pixel representations effectively, which validates our claim.


\noindent \textbf{Masking Ratio}.
Table~\ref{masking_ratio} illustrates the effect of varying masking ratios (\cf §\ref{subsec:strategical}). It depicts that a suitable masking ratio (50\% for ours) outperforms other settings with notable advantages. Such an empirical advantage can be explained by that the higher masking ratio may discard too much necessary information for self-learning paradigm via reconstruction to learn an effective image representation, whereas a low masking ratio may not be sufficient to increase the reconstruction difficulty and, consequently, the quality of predicting a flow map.




\subsection{Downstream Tasks}
\label{subsec:Down}

High-quality flow estimation plays a crucial role in many video-based downstream tasks. We show here quantitatively that TransFlow generalizes well and can help further improve the state-of-the-art of various video-based tasks, including video object detection, interpolation, and stabilization. 

\begin{table}[htb]
\centering
\scalebox{0.69}{\begin{tabular}{c|lcc}
\hline
\rowcolor{mygray}       
\cellcolor[HTML]{FFFFFF}                              & Method            & Backbone                        & \textit{mAP} (\%)             \\ \cline{2-4}   
\cellcolor[HTML]{FFFFFF}                          & RDN \cite{deng2019relation}               & ResNet-50                       & 76.7                             \\
\cellcolor[HTML]{FFFFFF}                         & RDN\textcolor{gray}{+ours}        & ResNet-50                       & 80.4 \textcolor{gray}{(3.7$\uparrow$)}    \\
\cellcolor[HTML]{FFFFFF}                                                                                             & SELSA \cite{wu2019sequence}            & ResNet-101                      & 80.3                             \\
\cellcolor[HTML]{FFFFFF}                                                                                             & SELSA\textcolor{gray}{+ours}       & ResNet-101                      & 82.9 \textcolor{gray}{(2.6$\uparrow$)}    \\
\cellcolor[HTML]{FFFFFF}                                                                                             & PTSEFormer \cite{wang2022ptseformer}        & ResNet-101                      & 87.4                             \\
\multirow{-7}{*}{\cellcolor[HTML]{FFFFFF}\textbf{\begin{tabular}[c]{@{}c@{}}Video\\ Object \\ Detection\end{tabular}}}                                               & PTSEFormer\textcolor{gray}{+ours}  & ResNet-101                      & 89.1 \textcolor{gray}{(1.7$\uparrow$)}    \\ \hline
& \cellcolor{mygray}Method & \cellcolor{mygray}PSNR       & \cellcolor{mygray}SSIM     \\ \cline{2-4} 
                                                                                             & SuperSloMo \cite{jiang2018super}       & 28.52                           & 0.891                            \\
                                                                                             & SuperSloMo\textcolor{gray}{+ours}  & 28.81 \textcolor{gray}{(0.29$\uparrow$)} & 0.905 \textcolor{gray}{(0.014$\uparrow$)}  \\
                                                                                             & IFR-Net \cite{kong2022ifrnet}          & 29.84                           & 0.920                            \\
\multirow{-5}{*}{\textbf{\begin{tabular}[c]{@{}c@{}}Video\\ Interpolation\end{tabular}}}                                                                                             & IFR-Net\textcolor{gray}{+ours}     & 30.02 \textcolor{gray}{(0.18$\uparrow$)}  & 0.932 \textcolor{gray}{(0.012$\uparrow$)}  \\ \hline
\multirow{5}{*}{\textbf{\begin{tabular}[c]{@{}c@{}}Video\\ Stabilization\end{tabular}}}     & \cellcolor{mygray}Method & \cellcolor{mygray}Distortion & \cellcolor{mygray}Stablity \\ \cline{2-4} 
                                                                                             & StabNet \cite{wang2018deep}          & 0.83                            & 0.75                             \\
                                                                                             & StabNet\textcolor{gray}{+ours}     & 0.85 \textcolor{gray}{(0.02$\uparrow$)}  & 0.79 \textcolor{gray}{(0.04$\uparrow$)}   \\
                                                                                             & PWStableNet \cite{zhao2020pwstablenet}       & 0.79                            & 0.80                             \\
                                                                                             & PWStableNet\textcolor{gray}{+ours} & 0.82 \textcolor{gray}{(0.03$\uparrow$)}  & 0.82 \textcolor{gray}{(0.02$\uparrow$)}   \\ \hline
\end{tabular}}
\vspace{-0.5mm}
\caption{\textbf{Quantitative comparison of downstream video task performance with our TransFlow.} See \S \ref{subsec:Down} for details.}
\label{downstream}
\end{table}

\noindent\textbf{Video Object Detection}.
We conduct our experiments on the ImageNet VID dataset \cite{russakovsky2015imagenet} containing over 1M frames for training and more than 100k frames for validation. As shown in Table~\ref{downstream}, adding TransFlow encoder feature in RDN \cite{deng2019relation}, SELSA \cite{wu2019sequence} and PTSEFormer  \cite{wang2022ptseformer} results in \textbf{3.7\%}, \textbf{2.6\%} and \textbf{1.7\%} improvement in the mean average precision (\textit{mAP}).




\vspace{0.5mm}
\noindent\textbf{Video Frame Interpolation}.
To evaluate our model for 8× interpolation, we train SuperSloMo \cite{jiang2018super} and IFR-Net \cite{kong2022ifrnet} on GoPro \cite{nah2017deep} training set with our TransFlow encoder features embedded, and test the trained model on GoPro testing set. As shown in Table~\ref{downstream}, the updated model outperform original methods with 2 input frames in both \textit{PSNR} and \textit{SSIM} (\ie \textbf{0.29} dB higher results than SuperSloMo and \textbf{0.18} dB higher than IFR-Net).  

\vspace{0.5mm}
\noindent\textbf{Video Stabilization}.
We follow the training configurations of StabNet \cite{wang2018deep} and PWStableNet\cite{zhao2020pwstablenet} and aggregate the learned features from the TransFlow encoder and the original encoder together for the later regressor. On the DeepStab \cite{wang2018deep} dataset which contains 61 pairs of stable and unstable videos, TransFlow feature-added method achieves a higher \textit{Distortion Value (D)} and \textit{Stability Score (S)} than the ones without it.

\vspace{-3mm}
\section{Conclusion}
\vspace{-1mm}
\label{sec:Concl}
We propose TransFlow, a pure transformer architecture for optical flow estimation. Extensive empirical analysis demonstrates that TransFlow establishes new records for public benchmarks. We trust that this work has the potential to provide valuable insights into the applicability of Transformer in more broad vision tasks.

\vspace{-3mm}
\section*{Acknowledgement}
\vspace{-2mm}
This work is supported by the National Science Foundation under Award No. 2242243.

{\small
\bibliographystyle{ieee_fullname}
\bibliography{egbib}

\begin{thebibliography}{10}\itemsep=-1pt

\bibitem{bao2021beit}
Hangbo Bao, Li Dong, and Furu Wei.
\newblock Beit: Bert pre-training of image transformers.
\newblock In {\em ICLR}, 2022.

\bibitem{black1993framework}
Michael~J Black and Padmanabhan Anandan.
\newblock A framework for the robust estimation of optical flow.
\newblock In {\em ICCV}, 1993.

\bibitem{bruhn2005lucas}
Andr{\'e}s Bruhn, Joachim Weickert, and Christoph Schn{\"o}rr.
\newblock Lucas/kanade meets horn/schunck: Combining local and global optic
  flow methods.
\newblock {\em IJCV}, 61(3):211--231, 2005.

\bibitem{butler2012naturalistic}
Daniel~J Butler, Jonas Wulff, Garrett~B Stanley, and Michael~J Black.
\newblock A naturalistic open source movie for optical flow evaluation.
\newblock In {\em ECCV}, 2012.

\bibitem{cao2022towards}
Zhiwen Cao, Dongfang Liu, Qifan Wang, and Yingjie Chen.
\newblock Towards unbiased label distribution learning for facial pose
  estimation using anisotropic spherical gaussian.
\newblock In {\em ECCV}, 2022.

\bibitem{carion2020end}
Nicolas Carion, Francisco Massa, Gabriel Synnaeve, Nicolas Usunier, Alexander
  Kirillov, and Sergey Zagoruyko.
\newblock End-to-end object detection with transformers.
\newblock In {\em ECCV}, 2020.

\bibitem{cheng2022physical}
Zhiyuan Cheng, James Liang, Hongjun Choi, Guanhong Tao, Zhiwen Cao, Dongfang
  Liu, and Xiangyu Zhang.
\newblock Physical attack on monocular depth estimation with optimal
  adversarial patches.
\newblock In {\em ECCV}, 2022.

\bibitem{chu2021conditional}
Xiangxiang Chu, Zhi Tian, Bo Zhang, Xinlong Wang, Xiaolin Wei, Huaxia Xia, and
  Chunhua Shen.
\newblock Conditional positional encodings for vision transformers.
\newblock {\em Arxiv preprint 2102.10882}, 2021.

\bibitem{cui2021tf}
Yiming Cui, Liqi Yan, Zhiwen Cao, and Dongfang Liu.
\newblock Tf-blender: Temporal feature blender for video object detection.
\newblock In {\em ICCV}, 2021.

\bibitem{dai2019transformer}
Zihang Dai, Zhilin Yang, Yiming Yang, Jaime Carbonell, Quoc~V Le, and Ruslan
  Salakhutdinov.
\newblock Transformer-xl: Attentive language models beyond a fixed-length
  context.
\newblock In {\em ACL}, 2019.

\bibitem{deng2019relation}
Jiajun Deng, Yingwei Pan, Ting Yao, Wengang Zhou, Houqiang Li, and Tao Mei.
\newblock Relation distillation networks for video object detection.
\newblock In {\em ICCV}, 2019.

\bibitem{devlin2018bert}
Jacob Devlin, Ming-Wei Chang, Kenton Lee, and Kristina Toutanova.
\newblock Bert: Pre-training of deep bidirectional transformers for language
  understanding.
\newblock In {\em NAACL}, 2019.

\bibitem{dosovitskiy2020image}
Alexey Dosovitskiy, Lucas Beyer, Alexander Kolesnikov, Dirk Weissenborn,
  Xiaohua Zhai, Thomas Unterthiner, Mostafa Dehghani, Matthias Minderer, Georg
  Heigold, Sylvain Gelly, et~al.
\newblock An image is worth 16x16 words: Transformers for image recognition at
  scale.
\newblock In {\em ICLR}, 2020.

\bibitem{dosovitskiy2015flownet}
Alexey Dosovitskiy, Philipp Fischer, Eddy Ilg, Philip Hausser, Caner Hazirbas,
  Vladimir Golkov, Patrick Van Der~Smagt, Daniel Cremers, and Thomas Brox.
\newblock Flownet: Learning optical flow with convolutional networks.
\newblock In {\em ICCV}, 2015.

\bibitem{feng2022disentangling}
Ziyue Feng, Liang Yang, Longlong Jing, Haiyan Wang, YingLi Tian, and Bing Li.
\newblock Disentangling object motion and occlusion for unsupervised
  multi-frame monocular depth.
\newblock In {\em ECCV}, 2022.

\bibitem{gehrig2021raft}
Mathias Gehrig, Mario Millh{\"a}usler, Daniel Gehrig, and Davide Scaramuzza.
\newblock E-raft: Dense optical flow from event cameras.
\newblock In {\em 3DV}, 2021.

\bibitem{he2022masked}
Kaiming He, Xinlei Chen, Saining Xie, Yanghao Li, Piotr Doll{\'a}r, and Ross
  Girshick.
\newblock Masked autoencoders are scalable vision learners.
\newblock In {\em CVPR}, 2022.

\bibitem{hofinger2020improving}
Markus Hofinger, Samuel~Rota Bul{\`o}, Lorenzo Porzi, Arno Knapitsch, Thomas
  Pock, and Peter Kontschieder.
\newblock Improving optical flow on a pyramid level.
\newblock In {\em ECCV}, 2020.

\bibitem{huang2022flowformer}
Zhaoyang Huang, Xiaoyu Shi, Chao Zhang, Qiang Wang, Ka~Chun Cheung, Hongwei
  Qin, Jifeng Dai, and Hongsheng Li.
\newblock Flowformer: A transformer architecture for optical flow.
\newblock In {\em ECCV}, 2022.

\bibitem{hui2018liteflownet}
Tak-Wai Hui, Xiaoou Tang, and Chen~Change Loy.
\newblock Liteflownet: A lightweight convolutional neural network for optical
  flow estimation.
\newblock In {\em CVPR}, 2018.

\bibitem{ilg2017flownet}
Eddy Ilg, Nikolaus Mayer, Tonmoy Saikia, Margret Keuper, Alexey Dosovitskiy,
  and Thomas Brox.
\newblock Flownet 2.0: Evolution of optical flow estimation with deep networks.
\newblock In {\em CVPR}, 2017.

\bibitem{jiang2018super}
Huaizu Jiang, Deqing Sun, Varun Jampani, Ming-Hsuan Yang, Erik Learned-Miller,
  and Jan Kautz.
\newblock Super slomo: High quality estimation of multiple intermediate frames
  for video interpolation.
\newblock In {\em CVPR}, 2018.

\bibitem{jiang2021learning}
Shihao Jiang, Dylan Campbell, Yao Lu, Hongdong Li, and Richard Hartley.
\newblock Learning to estimate hidden motions with global motion aggregation.
\newblock In {\em ICCV}, 2021.

\bibitem{jiang2021learningGMA}
Shihao Jiang, Dylan Campbell, Yao Lu, Hongdong Li, and Richard Hartley.
\newblock Learning to estimate hidden motions with global motion aggregation.
\newblock In {\em ICCV}, 2021.

\bibitem{jiang2021learningFEW}
Shihao Jiang, Yao Lu, Hongdong Li, and Richard Hartley.
\newblock Learning optical flow from a few matches.
\newblock In {\em CVPR}, 2021.

\bibitem{kondermann2016hci}
Daniel Kondermann, Rahul Nair, Katrin Honauer, Karsten Krispin, Jonas Andrulis,
  Alexander Brock, Burkhard Gussefeld, Mohsen Rahimimoghaddam, Sabine Hofmann,
  Claus Brenner, et~al.
\newblock The hci benchmark suite: Stereo and flow ground truth with
  uncertainties for urban autonomous driving.
\newblock In {\em CVPRW}, 2016.

\bibitem{kong2022ifrnet}
Lingtong Kong, Boyuan Jiang, Donghao Luo, Wenqing Chu, Xiaoming Huang, Ying
  Tai, Chengjie Wang, and Jie Yang.
\newblock Ifrnet: Intermediate feature refine network for efficient frame
  interpolation.
\newblock In {\em CVPR}, 2022.

\bibitem{li2022uniform}
Xiang Li, Wenhai Wang, Lingfeng Yang, and Jian Yang.
\newblock Uniform masking: Enabling mae pre-training for pyramid-based vision
  transformers with locality.
\newblock {\em arXiv preprint arXiv:2205.10063}, 2022.

\bibitem{liu2021fuseformer}
Rui Liu, Hanming Deng, Yangyi Huang, Xiaoyu Shi, Lewei Lu, Wenxiu Sun, Xiaogang
  Wang, Jifeng Dai, and Hongsheng Li.
\newblock Fuseformer: Fusing fine-grained information in transformers for video
  inpainting.
\newblock In {\em ICCV}, 2021.

\bibitem{luo2022learningKPA}
Ao Luo, Fan Yang, Xin Li, and Shuaicheng Liu.
\newblock Learning optical flow with kernel patch attention.
\newblock In {\em CVPR}, 2022.

\bibitem{luo2022learningAGflow}
Ao Luo, Fan Yang, Kunming Luo, Xin Li, Haoqiang Fan, and Shuaicheng Liu.
\newblock Learning optical flow with adaptive graph reasoning.
\newblock In {\em AAAI}, 2022.

\bibitem{luo2021upflow}
Kunming Luo, Chuan Wang, Shuaicheng Liu, Haoqiang Fan, Jue Wang, and Jian Sun.
\newblock Upflow: Upsampling pyramid for unsupervised optical flow learning.
\newblock In {\em CVPR}, 2021.

\bibitem{mayer2016large}
Nikolaus Mayer, Eddy Ilg, Philip Hausser, Philipp Fischer, Daniel Cremers,
  Alexey Dosovitskiy, and Thomas Brox.
\newblock A large dataset to train convolutional networks for disparity,
  optical flow, and scene flow estimation.
\newblock In {\em CVPR}, 2016.

\bibitem{menze2015joint}
Moritz Menze, Christian Heipke, and Andreas Geiger.
\newblock Joint 3d estimation of vehicles and scene flow.
\newblock {\em ISPRS Annals}, 2:427, 2015.

\bibitem{nah2017deep}
Seungjun Nah, Tae Hyun~Kim, and Kyoung Mu~Lee.
\newblock Deep multi-scale convolutional neural network for dynamic scene
  deblurring.
\newblock In {\em CVPR}, 2017.

\bibitem{prillo2020softsort}
Sebastian Prillo and Julian Eisenschlos.
\newblock Softsort: A continuous relaxation for the argsort operator.
\newblock In {\em ICML}, 2020.

\bibitem{russakovsky2015imagenet}
Olga Russakovsky, Jia Deng, Hao Su, Jonathan Krause, Sanjeev Satheesh, Sean Ma,
  Zhiheng Huang, Andrej Karpathy, Aditya Khosla, Michael Bernstein, et~al.
\newblock Imagenet large scale visual recognition challenge.
\newblock {\em IJCV}, 115(3):211--252, 2015.

\bibitem{sui2022craft}
Xiuchao Sui, Shaohua Li, Xue Geng, Yan Wu, Xinxing Xu, Yong Liu, Rick Goh, and
  Hongyuan Zhu.
\newblock Craft: Cross-attentional flow transformer for robust optical flow.
\newblock In {\em CVPR}, 2022.

\bibitem{sun2014quantitative}
Deqing Sun, Stefan Roth, and Michael~J Black.
\newblock A quantitative analysis of current practices in optical flow
  estimation and the principles behind them.
\newblock {\em IJCV}, 106(2):115--137, 2014.

\bibitem{sun2018pwc}
Deqing Sun, Xiaodong Yang, Ming-Yu Liu, and Jan Kautz.
\newblock Pwc-net: Cnns for optical flow using pyramid, warping, and cost
  volume.
\newblock In {\em CVPR}, 2018.

\bibitem{teed2020raft}
Zachary Teed and Jia Deng.
\newblock Raft: Recurrent all-pairs field transforms for optical flow.
\newblock In {\em ECCV}, 2020.

\bibitem{vaswani2017attention}
Ashish Vaswani, Noam Shazeer, Niki Parmar, Jakob Uszkoreit, Llion Jones,
  Aidan~N Gomez, {\L}ukasz Kaiser, and Illia Polosukhin.
\newblock Attention is all you need.
\newblock In {\em NIPS}, 2017.

\bibitem{wang2022ptseformer}
Han Wang, Jun Tang, Xiaodong Liu, Shanyan Guan, Rong Xie, and Li Song.
\newblock Ptseformer: Progressive temporal-spatial enhanced transformer towards
  video object detection.
\newblock In {\em ECCV}, 2022.

\bibitem{wang2020displacement}
Jianyuan Wang, Yiran Zhong, Yuchao Dai, Kaihao Zhang, Pan Ji, and Hongdong Li.
\newblock Displacement-invariant matching cost learning for accurate optical
  flow estimation.
\newblock In {\em NIPS}, 2020.

\bibitem{wang2018deep}
Miao Wang, Guo-Ye Yang, Jin-Kun Lin, Song-Hai Zhang, Ariel Shamir, Shao-Ping
  Lu, and Shi-Min Hu.
\newblock Deep online video stabilization with multi-grid warping
  transformation learning.
\newblock {\em TIP}, 28(5):2283--2292, 2018.

\bibitem{wang2020learning}
Qianqian Wang, Xiaowei Zhou, Bharath Hariharan, and Noah Snavely.
\newblock Learning feature descriptors using camera pose supervision.
\newblock In {\em ECCV}, 2020.

\bibitem{wang2022visual}
Wenguan Wang, Cheng Han, Tianfei Zhou, and Dongfang Liu.
\newblock Visual recognition with deep nearest centroids.
\newblock {\em NeurIPS}, 2022.

\bibitem{wang2022learning}
Wenguan Wang, James Liang, and Dongfang Liu.
\newblock Learning equivariant segmentation with instance-unique querying.
\newblock {\em NeurIPS}, 2022.

\bibitem{wang2022uformer}
Zhendong Wang, Xiaodong Cun, Jianmin Bao, Wengang Zhou, Jianzhuang Liu, and
  Houqiang Li.
\newblock Uformer: A general u-shaped transformer for image restoration.
\newblock In {\em CVPR}, 2022.

\bibitem{wu2019pay}
Felix Wu, Angela Fan, Alexei Baevski, Yann~N Dauphin, and Michael Auli.
\newblock Pay less attention with lightweight and dynamic convolutions.
\newblock In {\em ICLR}, 2019.

\bibitem{wu2019sequence}
Haiping Wu, Yuntao Chen, Naiyan Wang, and Zhaoxiang Zhang.
\newblock Sequence level semantics aggregation for video object detection.
\newblock In {\em ICCV}, 2019.

\bibitem{wu2021cvt}
Haiping Wu, Bin Xiao, Noel Codella, Mengchen Liu, Xiyang Dai, Lu Yuan, and Lei
  Zhang.
\newblock Cvt: Introducing convolutions to vision transformers.
\newblock In {\em ICCV}, 2021.

\bibitem{xu2021high}
Haofei Xu, Jiaolong Yang, Jianfei Cai, Juyong Zhang, and Xin Tong.
\newblock High-resolution optical flow from 1d attention and correlation.
\newblock In {\em ICCV}, 2021.

\bibitem{xu2020aanet}
Haofei Xu and Juyong Zhang.
\newblock Aanet: Adaptive aggregation network for efficient stereo matching.
\newblock In {\em CVPR}, 2020.

\bibitem{yang2022lite}
Chenglin Yang, Yilin Wang, Jianming Zhang, He Zhang, Zijun Wei, Zhe Lin, and
  Alan Yuille.
\newblock Lite vision transformer with enhanced self-attention.
\newblock In {\em CVPR}, 2022.

\bibitem{yin2019hierarchical}
Zhichao Yin, Trevor Darrell, and Fisher Yu.
\newblock Hierarchical discrete distribution decomposition for match density
  estimation.
\newblock In {\em CVPR}, 2019.

\bibitem{yuan2021tokens}
Li Yuan, Yunpeng Chen, Tao Wang, Weihao Yu, Yujun Shi, Zi-Hang Jiang,
  Francis~EH Tay, Jiashi Feng, and Shuicheng Yan.
\newblock Tokens-to-token vit: Training vision transformers from scratch on
  imagenet.
\newblock In {\em ICCV}, 2021.

\bibitem{zhang2021separable}
Feihu Zhang, Oliver~J Woodford, Victor~Adrian Prisacariu, and Philip~HS Torr.
\newblock Separable flow: Learning motion cost volumes for optical flow
  estimation.
\newblock In {\em ICCV}, 2021.

\bibitem{zhao2020pwstablenet}
Minda Zhao and Qiang Ling.
\newblock Pwstablenet: Learning pixel-wise warping maps for video
  stabilization.
\newblock {\em TIP}, 29:3582--3595, 2020.

\bibitem{zhao2020maskflownet}
Shengyu Zhao, Yilun Sheng, Yue Dong, Eric~I Chang, Yan Xu, et~al.
\newblock Maskflownet: Asymmetric feature matching with learnable occlusion
  mask.
\newblock In {\em CVPR}, 2020.

\end{thebibliography}
}

\end{document}